\documentclass[letterpaper, 10 pt, journal, twoside]{IEEEtran}

\usepackage{amsmath} 
\usepackage{amssymb}  

\usepackage{graphicx}
\usepackage{capt-of} 
\usepackage{overpic} 
\usepackage[export]{adjustbox} 
\usepackage{textcomp} 
\usepackage{tabularx}
\usepackage{colortbl} 
\usepackage{multirow}
\usepackage{booktabs}
\usepackage{diagbox} 
\usepackage{arydshln}
\usepackage{adjustbox} 
\usepackage{pgfplots} 
\usepackage{xcolor}

\usepackage{tikz}
\usetikzlibrary{arrows.meta, matrix}
\usepackage{forest}
\usepackage{siunitx}

\usepackage[caption=false, font=footnotesize]{subfig} 
\usepackage[hyperfootnotes=false]{hyperref}
\newcommand\scalemath[2]{\scalebox{#1}{\mbox{\ensuremath{\displaystyle #2}}}} 

\newcommand\copyrighttext{\footnotesize \textcopyright~2023 IEEE. Personal use of this material is permitted.  Permission from IEEE must be obtained for all other uses, in any current or future media, including reprinting/republishing this material for advertising or promotional purposes, creating new collective works, for resale or redistribution to servers or lists, or reuse of any copyrighted component of this work in other works.
}%

\newcommand\copyrightnotice{%
	\begin{tikzpicture}[remember picture,overlay]
	\node[anchor=south,xshift=0pt,yshift=2pt] at (current page.south) {\fbox{\parbox{\dimexpr\textwidth-\fboxsep-\fboxrule\relax}{\copyrighttext}}};
	\end{tikzpicture}%
}

\begin{document}

\title{SCENE: Reasoning about Traffic Scenes using\\Heterogeneous Graph Neural Networks}

\author{Thomas Monninger$^{\dagger, 1}$, Julian Schmidt$^{\dagger, 2, 3}$,\\Jan Rupprecht$^{2}$, David Raba$^{2}$, Julian Jordan$^{2}$,\\Daniel Frank$^{4}$, Steffen Staab$^{4, 5}$ and Klaus Dietmayer$^{3}$,~\IEEEmembership{Senior Member,~IEEE}%
\thanks{Manuscript received: August 26, 2022; Revised: November 21, 2022; Accepted: December 20, 2022.}
\thanks{This paper was recommended for publication by Editor Markus Vincze upon evaluation of the Associate Editor and Reviewers' comments.
This work was supported by the BMWK within the project "KI Delta Learning" (F\"orderkennzeichen 19A19013A) and the Deutsche Forschungsgemeinschaft (DFG, German Research Foundation) under Germany's Excellence Strategy - EXC 2075 – 390740016. \textit{(Corresponding author: Julian Schmidt)}} 
\thanks{$^{\dagger}$Thomas Monninger and Julian Schmidt are co-first authors. The order was determined alphabetically.}
\thanks{$^{1}$Thomas Monninger is with Mercedes-Benz R\&D North America, Sunnyvale, CA, USA (e-mail: thomas.monninger@mercedes-benz.com)}%
\thanks{$^{2}$Julian Schmidt, Jan Rupprecht, David Raba and Julian Jordan are with Mercedes-Benz AG, R\&D, Stuttgart, Germany (e-mail: \{julian.sj.schmidt, jan.rupprecht, david.raba, julian.jordan\}@mercedes-benz.com)}%
\thanks{$^{3}$Julian Schmidt and Klaus Dietmayer are with Ulm University, Institute of Measurement, Control and Microtechnology, Ulm, Germany (e-mail: klaus.dietmayer@uni-ulm.de)}%
\thanks{$^{4}$Daniel Frank and Steffen Staab are with University of Stuttgart, Institute of Parallel and Distributed Systems, Stuttgart, Germany (e-mail: \{daniel.frank, steffen.staab\}@ipvs.uni-stuttgart.de)}%
\thanks{$^{5}$Steffen Staab is with University of Southampton, Electronics and Computer Science, Southampton, United Kingdom}%
}

\markboth{IEEE Robotics and Automation Letters. Preprint Version. Accepted December, 2022}
{Monninger and Schmidt \MakeLowercase{\textit{et al.}}: SCENE} 

\maketitle

\begin{abstract}
Understanding traffic scenes requires considering heterogeneous information about dynamic agents and the static infrastructure.
In this work we propose SCENE, a methodology to encode diverse traffic scenes in heterogeneous graphs and to reason about these graphs using a heterogeneous Graph Neural Network encoder and task-specific decoders.
The heterogeneous graphs, whose structures are defined by an ontology, consist of different nodes with type-specific node features and different relations with type-specific edge features.
In order to exploit all the information given by these graphs, we propose to use cascaded layers of graph convolution.
The result is an encoding of the scene.
Task-specific decoders can be applied to predict desired attributes of the scene.
Extensive evaluation on two diverse binary node classification tasks show the main strength of this methodology:
despite being generic, it even manages to outperform task-specific baselines.
The further application of our methodology to the task of node classification in various knowledge graphs shows its transferability to other domains.
\end{abstract}


\IEEEpeerreviewmaketitle

\section{Introduction}
\IEEEPARstart{U}{nderstanding} traffic scenes is important for an autonomous vehicle such that it may develop a safe, effective and efficient plan of how to move forward.
For instance, whether a stationary car is parked or just temporarily stopped determines whether the autonomous vehicle should wait or overtake.
Understanding of traffic scenes requires reasoning about dynamic agents and static infrastructure in order to predict the intents of nearby dynamic agents (e.g., parked or temporarily stopped).
To this end, the vehicle needs to correctly estimate which sensory information is reliable and it must reason about the relative positions, features and trajectories of dynamic agents.
{\copyrightnotice}
Information about dynamic agents is conveyed by the perception systems of autonomous vehicles.
We raise the hypothesis that considering additional heterogeneous entities in a traffic scene might add valuable information.
In particular, reasoning should also involve knowledge about static infrastructure, which may either be perceived or in our case is provided by a High Definition (HD) map.
Thus, the problem of understanding traffic scenes boils down to integrating a plenitude of heterogeneous data, which may change over time, and reasoning about it in order to predict intents of nearby traffic agents.
This is difficult because data from heterogeneous sources may be structured in a myriad of ways and reasoning may require deriving complex relations and patterns across this heterogeneous data.

\begin{figure}[tpb]
	\vspace{0.06cm} 
	\centering
	\includegraphics[width=\columnwidth, trim=0.13cm 0cm 0cm 0cm, clip]{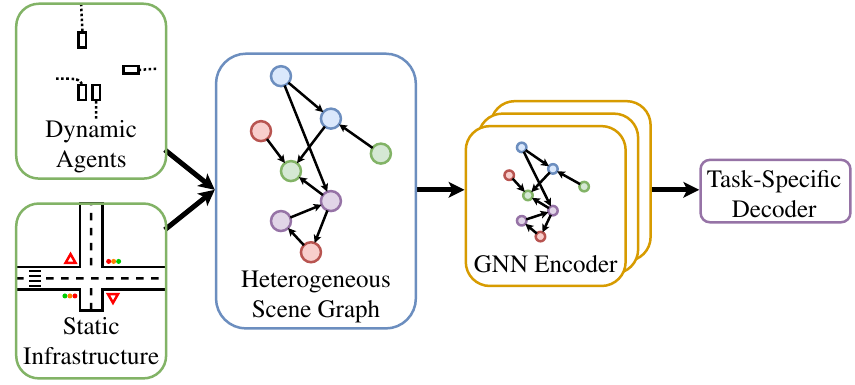}
	\caption{Overview of SCENE: The traffic scene is modeled in a heterogeneous scene graph with different node types and different relation types between these nodes. The combination of a generic GNN architecture, making use of cascaded layers of graph convolution, and a task-specific decoder is used to predict relevant information about the given scene.}
	\label{fig:motivation}
\end{figure}

Related work has tackled the problem of scene understanding from heterogeneous data by using machine learning approaches to reason about the scene.
Existing machine learning approaches that jointly leverage information about dynamic agents and static infrastructure, so far, have been based on rasterized representations (e.g., \cite{Hong2019}), have been handcrafted and task-specific (e.g., \cite{Behrendt2019}) or have been limited in their ability to consider heterogeneous data (e.g., \cite{Liang2020}).
Shortcomings of rasterized representations lie in the loss of information and task-specific approaches lack the ability to generalize to further tasks.\\

We propose SCENE \textit{(SCene Encoding NEtwork)}, a graph-based methodology to encode and perform reasoning about a traffic scene.
An overview of SCENE is given in Fig. \ref{fig:motivation}.
Inputs to SCENE are features provided by upstream perception components, which represent dynamic agents over a duration of \SI{3}{s}, as well as the abstract representation of static infrastructure, in our case given in an HD map.
This input is encoded in a heterogeneous scene graph with different node types and a set of typed relations between nodes.
In addition to this expressive representation, we provide the means for versatile reasoning and predictions in traffic scene graphs using a novel architecture based on Graph Neural Networks (GNNs).
With this work we contribute to the research on heterogeneous GNNs, defined by a recent survey as a core field of future work \cite{Wu2021}.
Our GNN model learns from examples how information about dynamic agents and static infrastructure of a traffic scene may be integrated and reasoned about, such that it can correctly predict unknown characteristics of entities or relations.
In order to show that this methodology is not task-specific, we evaluate it on two different binary node-classification tasks that correspond to the predictions (i) whether a car is parked or temporarily stopped and (ii) whether perceived information is reliable or not.

Our main contributions are:
\begin{itemize}
	\item We propose a novel way to model information about dynamic agents and static infrastructure of traffic scenes in one heterogeneous graph structure with edge features, allowing for an extensible and generic representation.
	\item We propose a novel GNN architecture that is able to perform reasoning on this heterogeneous graph.
	\item We extensively evaluate our proposed methodology on two diverse learning tasks.
	\item We quantify the effect of including heterogeneous data about additional scene entities and relations for those learning tasks in detailed ablation studies.
	\item We show that our GNN architecture transfers to applications beyond scene understanding, by applying it to the task of node classification in knowledge graphs.
\end{itemize}

\section{Related Work}
In this section, related work regarding reasoning about traffic scenes is discussed.
Existing approaches focus on the prediction of intents and trajectories of agents.

\subsection{Grid-based Approaches}
Grid-based approaches rasterize information in a bird's-eye view grid with multiple channels and use Convolutional Neural Networks (CNN) to learn from patterns in the given data in order to perform reasoning about the traffic scene.
One option is to use raw sensor data as input and project it into a bird's-eye view grid. \cite{Casas2018}.
Most recent approaches receive dynamic agents, extracted and processed from an upstream perception component, and information about the static infrastructure as input and render both into different channels of a grid \cite{Hong2019, Bansal2019, Chai2020, Djuric2020}.
Different to the heterogeneous graph from our work, a grid cannot represent complex relationships in an abstract form, e.g., the right of way between lanes \cite{Liang2020, Zeng2021}.

\subsection{Hybrid Approaches}
Hybrid approaches introduce a graph-based representation of the provided dynamic agents, but keep the grid-based representation for the static infrastructure.
The graph-based representation of agents allows for an agent-wise encoding, considering semantic attributes and temporal information.
Reasoning on these encodings is done via GNNs \cite{Salzmann2020, Tang2019, Gilles2021}, which can consider edges in the graph to derive complex interaction patterns between agents.
In contrast to our work, these hybrid approaches still come with the aforementioned limitations of not representing complex relationships that involve static infrastructure.

\subsection{Graph-based Approaches}
Graph-based approaches model both, dynamic agents and static infrastructure via graph structures.
This idea of holistically modeling scenes in a graph structure and reasoning on it originated from the field of image retrieval \cite{Johnson2015}.

Since graph-based approaches work on sparse graphs instead of dense grids, these approaches tend to be more memory efficient \cite{Khandelwal2020_ARXIV}.
Analogously to the hybrid approaches, GNNs are used to model interactions between dynamic agents.

Early work of Ulbrich et al. \cite{Ulbrich2014} proposes an ontology for representing a scene graph for autonomous driving, but do not provide means for reasoning on the graph.
Tian et al. \cite{Tian2020_ARXIV} propose a simplified approach by not modeling lanes explicitly.
This is a limitation compared to our work because their approach cannot capture topological nor regulatory relationships between lanes.

Gao et al. \cite{Gao2020} propose VectorNet, which shares the concept of creating one global graph and serves as a baseline in our evaluations.
In contrast, their graph is homogeneous and fully-connected.
The homogeneous representation is obtained by learning a node embedding for each of the heterogeneous entities in the scene, including dynamic agents and static infrastructure (e.g., crosswalks and lanes).
Their fully-connected graph has only one type of edges, which requires the network to implicitly learn different semantic relations between nodes based on their embeddings.
As a drawback, their representation does not capture edge features.
However, edge features enable the inclusion of additional relational information in the graph, which we evaluate as advantageous in our ablation study.
The use of edge features in general is very limited in recent publications.
Approaches either only use spatial relations as edge features (e.g., distances or headings between dynamic agents) \cite{Ma2019, Li2020a_ARXIV_IS_EQUAL_Li2021_ARXIV}, or intermediate representations by combining information of two connected nodes \cite{Li2020, Hu2022}.

Li et al. \cite{Li2020a} use one graph to model the interactions between an ego vehicle and its nearby vehicles (ego-thing graph) and one graph to model interactions between the ego vehicle and its static infrastructure (ego-stuff graph).
For the ego-stuff graph, only graph edges between the node of the ego vehicle and stuff nodes are allowed.
This modeling limits their reasoning to the ego vehicle only, while our approach is capable of also reasoning over patterns between non-ego vehicles.
Kumar et al. \cite{Kumar2021} lift that restriction and create a heterogeneous graph in which all agents are fully connected to each other as well as to nodes of the static infrastructure within a fixed radius.
Still, no relations between the static infrastructure are explicitly modeled in the graph.
Both approaches suffer from the aforementioned limitation that the graph contains no explicitly modeled edge features between entities of the static infrastructure.

Other approaches \cite{Liang2020, Zeng2021, Khandelwal2020_ARXIV} explicitly model the lane topology in a graph in order to incorporate knowledge about the static infrastructure.
These approaches utilize specialized mechanisms in the inference process to include lane information, allowing a transductive exchange of information between agents via the underlying lanes.
However, they do not cover other entities of the static infrastructure, such as crosswalks or traffic lights.
In contrast, our methodology is generic and freely extensible in a sense that it can capture various information in a heterogeneous graph by using typed nodes.
Furthermore, our methodology allows for modeling relations with arbitrary type and edge features between these nodes, which we demonstrate to be a valuable addition.
The result is one heterogeneous graph that explicitly models all aspects of a given traffic scene without limitations to specific use cases.

\section{Methodology}
This section describes our proposed methodology.
Firstly, we define a graph ontology to model the given dynamic agents and static infrastructure in a heterogeneous scene graph.
Secondly, we use a learning-based approach to predict relevant information from this scene graph.

\subsection{Heterogeneous Scene Graph Ontology}
We represent a scene by a directed heterogeneous graph $\mathcal{G} = ( \mathcal{V}, \mathcal{E}, \mathcal{T}, \mathcal{R}, \phi ) $.
Every node $v_i \in \mathcal{V}$ has a feature vector $\mathbf{v}_i$.
The edge $e_{j,r,i} = (v_j, r, v_i) \in \mathcal{E}$ between the source node $v_j$ and the destination node $v_i$ with the relation type $r \in \mathcal{R}$ has a feature vector $\mathbf{e}_{j,r,i}$.
The type of node $v$ is defined by the type operator $\phi: \mathcal{V} \to \mathcal{T}$, with $\mathcal{T}$ being the set of allowed node types.
We define the domain type operator $dom: \mathcal{R} \to \mathcal{T}$ and range type operator $ran: \mathcal{R} \to \mathcal{T}$ to map a relation type $r$ to the source and target node types, respectively.
Each relation type $r$ has a fixed source and destination node type: $\forall (v_j,r,v_i) \quad \phi(v_j) \in dom(r) \ \text{and} \ \phi(v_i) \in ran(r)$.

\begin{figure}[thbp]
	\centering
    \vspace{-0.2cm}
	\begin{tikzpicture}
	\begin{scope}[every node/.style={circle,thick,draw, font=\footnotesize, minimum width=38pt}]
	\node (crosswalk) at (0,2) {crosswalk};
	\node (light) at (3.5,3.5) {light};
	\node (lane) at (3.5,0.5) {lane};
	\node (agent) at (7,2) {agent};
	\node (stop) at (7,-1) {stop};
	\end{scope}
	\begin{scope}[>={Stealth[black]},
	every node/.style={fill=white, font=\footnotesize},
	every edge/.style={draw=red, very thick, inner sep=1pt}] 
	\path[->] (agent) edge[draw=pink, loop, out=90+19, in=90-19, min distance=20mm] node {interacts} (agent);
	\path [->] (agent) edge[draw=blue, bend right=20] node {on} (lane);
	\path [->] (lane) edge[draw=red, bend right=20] node {under} (agent);
	\path[->] (lane) edge[draw=blue, loop, out=180+38, in=180, min distance=20mm] node {conflict} (lane);
	\path[->] (lane) edge[draw=blue, loop, out=180+38+10+38, in=180+38+10, min distance=20mm] node {connection} (lane);
	\path[->] (lane) edge[draw=blue, loop, out=180+38+10+38+10+38, in=180+38+10+38+10, min distance=20mm] node {precedence} (lane);
	\path [->] (crosswalk) edge[draw=blue] node {overlaps} (lane);
	\path [->] (light) edge[draw=blue] node {controls} (lane);
	\path [->] (agent) edge[draw=green, right, bend right=10] node {crosses} (crosswalk);
	\path [->] (light) edge[draw=green] node {signals} (crosswalk);
	\path [->] (stop) edge[draw=blue] node {stops} (lane);
	\end{scope}
	\end{tikzpicture}
    \vspace{-0.5cm}
	\caption{Node types and allowed relation types of the proposed ontology. Different colors are used to indicate the order of our proposed flow of information.}
	\label{fig:ontology}
\end{figure}
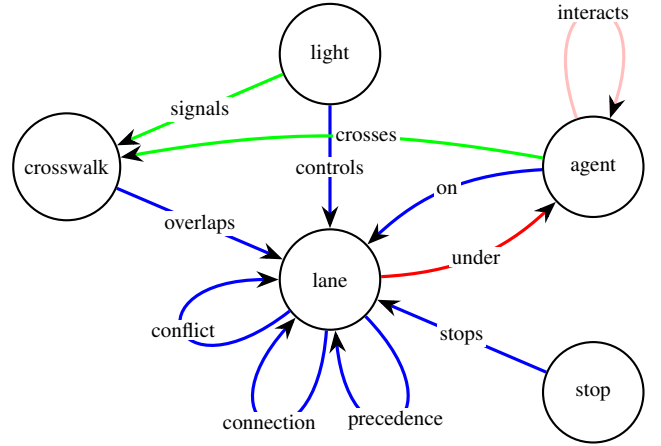

Fig. \ref{fig:ontology} illustrates our used node types and our used relation types between these node types.
Each node and edge has a corresponding feature vector.
The relation types belong to three groups.
Relations between agents are of type \textit{interacts}.
They are dynamically generated for each pair based on the assumption that all agents can interact with each other.
Relations between agents and map entities are of the types \textit{on}, \textit{under} and \textit{crosses}.
All valid relations are dynamically derived from the geometric constellation.
The remaining relation types link map entities and are given by the HD map.

\subsection{Reasoning on the Heterogeneous Scene Graph}
Reasoning on the generated heterogeneous scene graph is done with the encoder-decoder architecture presented below.
The encoder first aggregates information of a traffic scene into embeddings of the \textit{agent} nodes.
From these embeddings, task-specific decoders can directly predict agent-specific attributes (e.g., intents or trajectories).
Encoder and decoder are jointly trained with task-specific data in a supervised manner.
The focus of this work is the generic encoder.

\subsubsection{Encoder}
Multiple layers of graph convolution are cascaded to aggregate information regarding the heterogeneous scene.
Thanks to their invariance properties \cite{Keriven2019}, graph convolutional layers can learn general, abstract patterns from concrete scenes.
We show that this principle works for different classification tasks.

For graph convolution, a variety of operators is applicable.
We follow the principle described in \cite{Wang2021}, allowing to incorporate edge features in the established Graph Attention Network (GAT) operator \cite{Velickovic2018}.
For better error propagation and to avoid over-smoothing, we add a residual connection for $\mathbf{\Theta}_{\mathrm{s}, r} \cdot \mathbf{v}_i$ to the operator.
The update of node $v_i$ under consideration of neighboring nodes connected via the relation type $r$ is given by
\begin{equation}
\begin{aligned}
\mathbf{v}_{i, r}^\prime = {} & \mathrm{EdgeGAT}_r( \mathbf{v}_i ) = \\ &
\mathbf{\Theta}_{\mathrm{s}, r} \cdot \mathbf{v}_i + \\ &
\Big{\|}_{k=1}^K \left( \sum\limits_{j \in \mathcal{N}_r(v_i)} \alpha^k_{j, r, i} \left( \mathbf{\Theta}^k_{\mathrm{n}, r} \cdot \mathbf{v}_j + \mathbf{\Theta}^k_{\mathrm{e}, r} \cdot \mathbf{e}_{j,r,i} \right) \right) \text{.}
\end{aligned}
\end{equation}
$\mathbf{\Theta}$ is used to denote learnable weight matrices for the transformation of features of the node to update (s=self), neighboring nodes (n=neighbor) and edge features (e=edge).
$K$ corresponds to the number of attention heads and ${\|}$ denotes the concatenation operator.
Attention weights are obtained by
\begin{equation}
\begin{aligned}
\alpha^k_{j, r, i} = {} & \mathrm{softmax}_{r, i} \Big( \mathrm{LeakyReLU} \big( \\ & {\mathbf{a}_r^k}^T [ \mathbf{\Theta}^k_{\mathrm{n}, r} \cdot \mathbf{v}_i || \mathbf{\Theta}^k_{\mathrm{n}, r} \cdot \mathbf{v}_j || \mathbf{\Theta}^k_{\mathrm{e}, r} \cdot \mathbf{e}_{j,r,i} ] \big) \Big) \text{,}
\end{aligned}
\end{equation}
with $\mathbf{a}$ corresponding to a learnable vector.
$\mathrm{softmax_{r, i}}$ stands for the normalization by all incoming edges of node $i$ connected via relation type $r$.

While EdgeGAT is able to aggregate information of one specific relation type, reasoning on a heterogeneous graph requires aggregating information of neighboring nodes that are possibly connected via different relation types.
Adapted from Schlichtkrull et al. \cite{Schlichtkrull2018}, we define the node update of one heterogeneous GNN layer as
\begin{equation}
\mathbf{v}_{i}^\prime = \mathrm{ReLU} \left( \sum\limits_{r \in \mathcal{R}} \mathbf{v}_{i,r}^\prime \right) \text{.}
\end{equation}
We denote the combination of EdgeGAT and the aggregation of the resulting embeddings over multiple relation types as HetEdgeGAT.

We propose to use cascaded layers of HetEdgeGAT in order to aggregate information of the scene into \textit{agent} nodes.
The flow of information towards the \textit{agent} nodes is represented by the color of the relation types in Fig. \ref{fig:ontology}:
The first layer aggregates information into \textit{crosswalk} nodes (green).
Subsequent layers aggregate information into \textit{lane} nodes (blue) and \textit{agent} nodes (red).
The last layer of graph convolution then considers social interaction between agents and updates the \textit{agent} nodes again (pink).

\subsubsection{Decoder}
The decoder is task-specific.
We use a Multilayer Perceptron (MLP) and apply it to the encodings of \textit{agent} nodes for the two binary node classification tasks considered in the experiments section.

\section{Experiments} \label{sec:experiments}
In this section we describe the extensive evaluation of our proposed methodology.

\subsection{Learning Tasks}
We evaluate our methodology on two diverse binary node classification tasks:
\begin{enumerate}
	\item Classification whether an agent is parked or not. We consider one prior publication introducing and addressing this task as a baseline \cite{Behrendt2019}.
	\item Classification whether an agent is a ghost or not. Ghosts are unreliable detections of agents by upstream perception components that do not exist in real world, i.e., false positive detections. We are not aware of prior publications that consider static infrastructure for this task. Comparison is done with an approach that considers only dynamic information \cite{Aeberhard2011}.
\end{enumerate}

\subsection{Dataset}
Experiments are carried out on a large-scale, in-house dataset with over $22\,400$ sequences, each coming with \SI{3}{s} of temporal history.
They are extracted from in-vehicle recordings from different areas in Germany and the U.S. with a sampling rate of \SI{10}{Hz}.
Camera, LiDAR and Radar detections are fused together in order to detect surrounding dynamic agents.
The dataset includes diverse environments (e.g., urban, rural and highway) as well as diverse scenarios (e.g., driving and yielding).
To the best of our knowledge, there is no publicly available dataset for scene understanding that similarly provides manually annotated semantic attributes for dynamic agents in combination with an extensively attributed heterogeneous HD map.

By deriving correspondences between manually annotated agents and agent detections from upstream perception components, more than $430\,000$ labels per training tasks are generated.
For our experiments, we use a dataset split of $60\%$ (training), $30\%$ (validation) and $10\%$ (testing).

\subsection{Model Implementation Details}
We use an extensive set of features for nodes and edges in the scene graph to explicitly model all available knowledge of the scene.
Features are provided by the upstream perception components and the HD map.
Table \ref{tab:node_features} and Table \ref{tab:edge_features} list the used feature sets for each node type and each relation type.
To propagate uncertainty of the perception component to the model, the feature vectors of \textit{agent} nodes contain covariances and confidence values.
All features that express a category type are one-hot encoded.
While this set of features is given by our perception and HD map, the proposed methodology can use any arbitrary set of features.

\begin{figure*}[thpb]
	\centering
	\includegraphics[width=\textwidth, trim=0cm 0cm 0cm 0cm, clip]{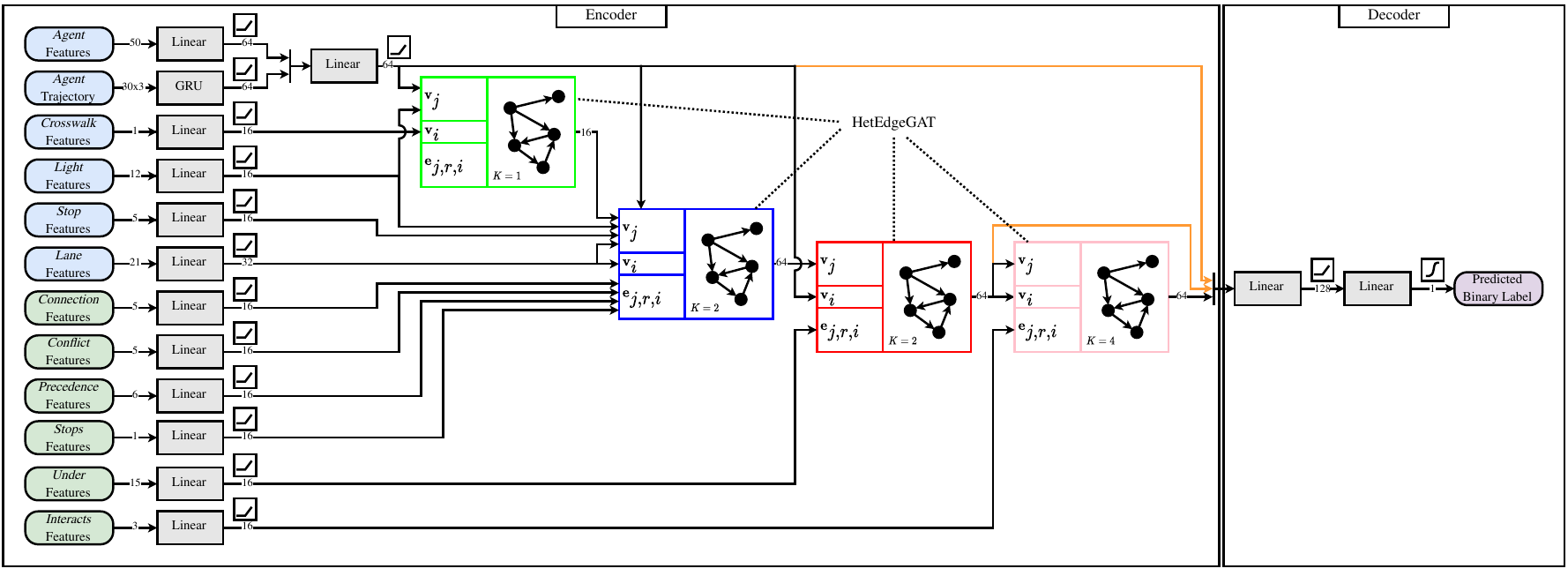}
	\vspace{-0.6cm}
	\caption{Implementation details of the SCENE encoder and the task-specific MLP-decoder: Four cascaded layers of HetEdgeGAT (green, blue, red, pink) are used to combine information of nodes of different types and their relations in order to update the feature vector of \textit{agent} nodes. Residual connections (orange) prevent over-smoothing. For the two binary node classification tasks, an MLP is then used to generate a classification score.}
	\label{fig:implementation_details}
	\vspace{-0.1cm}
\end{figure*}

\begin{table}[!t]
	\scriptsize
	\newcommand{\tabitem}{\hangindent=0.21cm ~\llap{\textbullet}~}
	\caption{Node type-specific features}
	\vspace{-0.1cm}
	\label{tab:node_features}
	\centering
	\begin{tabularx}{1\columnwidth}{l >{\raggedright\arraybackslash}X}
		\toprule
		Node type & Features                                                                                                                         \\ \midrule
		agent     &  \tabitem State vector (position, velocity, acceleration, yaw, yaw rate) with corresponding covariances                                         \\
		& \tabitem Tracking properties (e.g., max. velocity, tracked time)                                      \\
		& \tabitem Bounding box dimensions                                                                                                 \\
		& \tabitem Estimate of agent type (e.g., car, truck, two-wheeler)                                                                \\
		& \tabitem Sensor specific detection and existence probabilities \\
		& \tabitem Existence confidence calculated according to \cite{Aeberhard2011}                                                                                         \\
		& \tabitem Trajectory of the past three seconds as a series of positional and angular differences                                  \\  \arrayrulecolor{black}\specialrule{0.2pt}{0.4pt}{1.6pt}
		lane      & \tabitem Type (car, bike, shoulder, parking)                                                                                     \\
		& \tabitem Geometric properties (length, min. and max. width, max. curvature)                                                                                     \\
		& \tabitem Maximum legal speed                                                                                                           \\
		& \tabitem Left and right boundary types                                                        \\
		& \tabitem Turn type                                                                                 \\ \arrayrulecolor{black}\specialrule{0.2pt}{0.4pt}{1.6pt}
		crosswalk  & \tabitem Is signaled                                                                                                        \\ \arrayrulecolor{black}\specialrule{0.2pt}{0.4pt}{1.6pt}
		stop      & \tabitem Type (e.g., stop, crosswalk, yield)                                                                           \\ \arrayrulecolor{black}\specialrule{0.2pt}{0.4pt}{1.6pt}
		light     &  \tabitem Type (e.g., car, pedestrian)                                                                                    \\
		& \tabitem State (e.g., red, yellow) \\
		& \tabitem Is deactivatable                                                                                                           \\ \bottomrule
	\end{tabularx}
	\vspace{-0.1cm}
\end{table}

\begin{table}[!t]
	\scriptsize
	\newcommand{\tabitem}{\hangindent=0.21cm ~\llap{\textbullet}~}
	\caption{Relation type-specific edge features}
	\vspace{-0.1cm}
	\label{tab:edge_features}
	\centering
	\begin{tabularx}{1\columnwidth}{l >{\raggedright\arraybackslash}X}
		\toprule
		Relation type & Features                                                                 \\ \midrule
		interacts     & \tabitem Geometric differences (position, velocity, angle)            \\ \arrayrulecolor{black}\specialrule{0.2pt}{0.4pt}{1.6pt}
		under         & \tabitem Assignment probability                                          \\
		              & \tabitem Frenet state (position, velocity) at agent position                                   \\
		              & \tabitem Lane properties at agent position                                \\
		              & \tabitem Gap to lane boundaries at agent position                         \\
		              & \tabitem Behavior primitive of agent in lane (e.g., following, crossing) \\ \arrayrulecolor{black}\specialrule{0.2pt}{0.4pt}{1.6pt}
		connection    & \tabitem Type (e.g., precede, left neighbor)                             \\ \arrayrulecolor{black}\specialrule{0.2pt}{0.4pt}{1.6pt}
		conflict      & \tabitem Type (e.g., cross, merge)            \\ \arrayrulecolor{black}\specialrule{0.2pt}{0.4pt}{1.6pt}
		precedence    & \tabitem Type (e.g., higher, lower)        \\ \arrayrulecolor{black}\specialrule{0.2pt}{0.4pt}{1.6pt}
				stops    & \tabitem Longitudinal position in lane                            \\ \bottomrule
	\end{tabularx}
	\vspace{-0.2cm}
\end{table}

Details of the final implementation are shown in Fig. \ref{fig:implementation_details}, including the dimensions of all feature vectors.
The features of all nodes and edges are type-specifically encoded with a single linear layer and $\mathrm{ReLU}$.
Static and temporal aspects of agents are encoded separately and concatenated thereafter.
The static encoding uses a linear layer with $\mathrm{ReLU}$.
The temporal encoding uses a Gated Recurrent Unit (GRU) and $\mathrm{ReLU}$ to process the trajectory feature.
Following the idea of \cite{Liang2020}, the trajectory feature contains a fixed-length series of positional and angular differences of the last $30$ timesteps (\SI{3}{s}).
We add a binary flag that indicates whether an entry contains a valid measurement for each timestep.

In order to avoid over-smoothing, which is one of the main issues of multilayer GNNs \cite{Zhou2020}, we exploit two concatenated residual connections (orange).
These allow the decoder to combine high and low-level features of \textit{agent} nodes.

Binary cross-entropy is used as a loss function.
The model is trained with Adam optimizer \cite{Kingma2015} with a learning rate of $10^{-5}$ and a batch size of $32$.
Dropout with a rate of $0.3$ is used for the two linear layers of the decoder.

\subsection{Baselines}
For both tasks, we compare our generic methodology to multiple task-specific and generic baselines.

\subsubsection{Task-specific Baselines for the Parked Attribute}
The \textbf{velocity baseline} evaluates the velocity of each car in a given scene.
Stationary cars (zero velocity) are labeled as parked and vice versa.

The \textbf{logistic regression baseline} uses a handcrafted set of features based on the inputs provided by upstream perception components and the HD map.
This baseline has been specifically designed for classifying the parked attribute.

We also consider two approaches of one prior publication addressing parked car classification \cite{Behrendt2019}, namely the heuristic approach and the MLP approach, which operates on only three features for each agent.
Both approaches utilize features that contain information about the agent and one underlying lane.
We call these baselines \textbf{heuristic} and \textbf{MoveMLP3}.

\subsubsection{Task-specific Baseline for the Ghost Attribute}
The \textbf{existence confidence baseline} gives an estimate about the existence of an agent.
This baseline approach \cite{Aeberhard2011} uses a method based on Dempster-Shafer evidence theory to estimate a fused existence confidence about an agent based on detections from multiple sensor modalities.
Note that the resulting existence confidence is also part of the input features of \textit{agent} nodes.
A comparison to this baseline therefore shows the benefit of additionally considering social context and map context.

\subsubsection{Generic Baselines}
The \textbf{MLP baseline} contains four linear layers with $\mathrm{ReLU}$ between these layers.
It operates directly on the features of \textit{agent} nodes and does not process the graph structure.
In comparison to our proposed model, this baseline shows the effect of neglecting relational information about social context, defined by nearby dynamic agents, and map context, defined by the static infrastructure.

The \textbf{R-GCN baseline} applies the Relational Graph Convolutional Network \cite{Schlichtkrull2018}, typically used as a common approach to reason about knowledge graphs, to our scene graph.
We extend the original R-GCN approach by introducing edge features, which allows our implementation of R-GCN to use the same input features as SCENE.
After four layers of R-GCN, an MLP decoder is used to predict the labels.
Feature vector sizes of nodes and edges are similar to the ones used in SCENE.

To compare our methodology to the current-state-of-the-art in scene encoding, we adapt VectorNet \cite{Gao2020} to our input representation.
The \textbf{VectorNet-like baselines} therefore rely on learning from a fully-connected, homogeneous graph.
In contrast to SCENE, the features of all heterogeneous nodes are type-specificially encoded into a joint feature space with size $64$ to get a homogeneous graph.
The vanilla VectorNet-like baseline does not use edge features.
In order to allow a fair comparison to SCENE, we also extend the original VectorNet approach by the introduction of edge features.
All existing edges are type-specifically encoded to a size of $16$.
To obtain the required fully-connected graph, edges are instantiated with a zero vector of size $16$ between all nodes that are not yet connected.
A binary flag concatenated to the edge feature vector is used to indicate whether an edge is valid ($1$) or invalid ($0$).
Despite this leading to a fully-connected, homogeneous graph, connectivity information of our initial scene graph is still conserved in the edge features.
One layer of EdgeGAT is used on the resulting fully-connected graph.
Similar to SCENE, the labels are then predicted with an MLP decoder.

\subsection{Metrics}
F-Score (F\textsubscript{1}) and accuracy (Acc) are used for evaluation.

\begin{table}[!t]
	\scriptsize
	\caption{Results on the test set}
	\vspace{-0.1cm}
	\label{tab:results_test}
	\setlength{\tabcolsep}{1pt}
	\centering
	\begin{tabularx}{\columnwidth}{Xllll}
		\toprule
		\multirow{2}{*}{Method}                      &                              \multicolumn{2}{c}{Parked}                               &                               \multicolumn{2}{c}{Ghost}                               \\
		\cmidrule(lr){2-3}\cmidrule(lr){4-5}         & F\textsubscript{1} (\%)                   & Acc (\%)                                  & F\textsubscript{1} (\%)                   & Acc (\%)                                  \\ \midrule
		Velocity                                     & $76.51$                                   & $81.41$                                   & -                                         & -                                         \\
		Logistic regression                          & $89.79$                                   & $93.21$                                   & -                                         & -                                         \\
		Heuristic \cite{Behrendt2019}                & $86.75$                                   & $90.47$                                   & -                                         & -                                         \\
		MoveMLP3 \cite{Behrendt2019}                 & $88.56\scalemath{0.7}{\pm 0.15}$          & $92.78\scalemath{0.7}{\pm 0.07}$          & -                                         & -                                         \\ \midrule
		Existence confidence \cite{Aeberhard2011}    & -                                         & -                                         & $53.48$                                   & $66.01$                                   \\ \midrule
		Naive prior                                  & $0.00$                                    & $67.33$                                   & $66.88$                                   & $50.24$                                   \\
		MLP                                          & $75.79\scalemath{0.7}{\pm 0.89}$          & $83.23\scalemath{0.7}{\pm 0.40}$          & $79.93\scalemath{0.7}{\pm 0.67}$          & $80.73\scalemath{0.7}{\pm 0.44}$          \\
		R-GCN \cite{Schlichtkrull2018}               & $89.68\scalemath{0.7}{\pm 0.81}$          & $93.11\scalemath{0.7}{\pm 0.58}$          & $78.76\scalemath{0.7}{\pm 0.36}$          & $79.87\scalemath{0.7}{\pm 0.24}$          \\
		VectorNet-like \cite{Gao2020}                & $73.90\scalemath{0.7}{\pm 1.43}$          & $82.71\scalemath{0.7}{\pm 0.74}$          & $74.63\scalemath{0.7}{\pm 4.17}$          & $75.84\scalemath{0.7}{\pm 1.68}$          \\
		VectorNet-like (w/ edge feat) \cite{Gao2020} & $89.18\scalemath{0.7}{\pm 1.28}$          & $93.08\scalemath{0.7}{\pm 0.75}$          & $80.93\scalemath{0.7}{\pm 1.26}$          & $81.40\scalemath{0.7}{\pm 0.98}$          \\ \midrule
		Ours (using HetEdgeGAT)\footnotemark         & $\mathbf{91.17}\scalemath{0.7}{\pm 0.71}$ & $\mathbf{94.29}\scalemath{0.7}{\pm 0.46}$ & $80.56\scalemath{0.7}{\pm 0.77}$          & $81.44\scalemath{0.7}{\pm 0.71}$          \\
		Ours (using HetEdgeGatedGCN)                 & $90.09\scalemath{0.7}{\pm 0.29}$          & $93.54\scalemath{0.7}{\pm 0.15}$          & $\mathbf{82.42}\scalemath{0.7}{\pm 1.33}$ & $\mathbf{82.83}\scalemath{0.7}{\pm 1.07}$ \\
		Ours (using HetEdgeSAGE)                     & $91.11\scalemath{0.7}{\pm 0.43}$          & $94.19\scalemath{0.7}{\pm 0.34}$          & $80.72\scalemath{0.7}{\pm 1.23}$          & $81.68\scalemath{0.7}{\pm 0.71}$          \\
		Ours (using HetEdgeGAT)$^*$                  & $90.16\scalemath{0.7}{\pm 1.42}$          & $93.49\scalemath{0.7}{\pm 1.10}$          & $81.05\scalemath{0.7}{\pm 1.26}$          & $81.93\scalemath{0.7}{\pm 0.80}$          \\ \bottomrule
	\end{tabularx}
	\begin{flushleft}
	\footnotemark[1]Selected for all further experiments.\quad $^*$Multi-task training.
	\end{flushleft}
	\vspace{-0.3cm}
\end{table}

\subsection{Quantitative Results}
The models were trained over five random seeds to minimize stochasticity in the results.
The resulting average and standard deviation of the performance metrics on the test split are shown in Table \ref{tab:results_test}.
Besides GAT, we evaluated our approach using different operators for graph convolution, including variants of Gated Graph Convolutional Neural Networks (EdgeGatedGCN) \cite{Bresson2018_ARXIV} and GraphSAGE (EdgeSAGE) \cite{Hamilton2017}.
They all consistently perform well, which suggests that the graph convolution operator is interchangeable, also with regards to the architecture.
Therefore, our methodology can benefit from upcoming advances in the field of GNNs.
HetEdgeGAT was selected for all further experiments because it uses the least number of parameters.
Also, Schmidt et al. \cite{Schmidt2022} show that the resulting attention weights of \textit{interacts} relations offer additional interpretability, as they are a direct measure for interactions.

Comparing the results of our methodology with the MLP baseline supports our initial hypothesis that our proposed way of modeling a scene in a heterogeneous scene graph adds valuable information.

Our generic methodology outperforms all task-specific and generic baselines on both tasks.
The last row in Table \ref{tab:results_test} shows the result of the model simultaneously trained on both learning tasks.
The performance is on par with the single-task setup, which supports the indication that our method works as a generic scene encoder.
The VectorNet-like baseline extended with edge features performs much better than the vanilla VectorNet-like baseline, which supports the intuition that adding relational attributes provides valuable information for scene understanding.
Comparing the average number of Floating-Point Operations (FLOPs) of the VectorNet-like baseline with edge features ($6.24\cdot10^8$ FLOPs) and our SCENE approach ($\mathbf{4.57\cdot10^7}$ FLOPs) shows that our approach has more than an order of magnitude less computational complexity.
The higher complexity of the VectorNet-like baselines comes from applying convolution over the fully-connected graph, which also results in significantly higher GPU memory requirements and training time than our approach.

\begin{figure*}[thpb]
	\centering
	\subfloat{%
		\begin{overpic}[width=0.32\textwidth, trim=0cm 0cm 0cm 3.3cm, clip, frame]{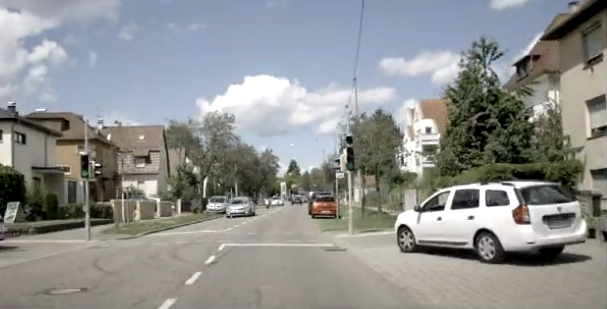}
			\setlength\fboxsep{0pt}
			\tiny
			\put(1, 25.2){\pgfsetfillopacity{0.8}\colorbox{white}{\framebox(27,4){\pgfsetfillopacity{1}Parked classification}}}
	\end{overpic}}
	\hfill
	\subfloat{%
		\begin{overpic}[width=0.32\textwidth, trim=0cm 0cm 0cm 3cm, clip, frame]{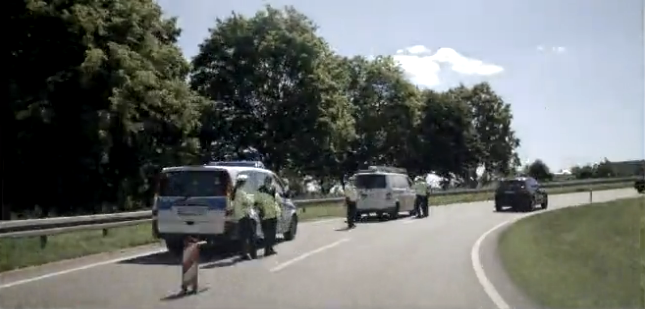}
			\setlength\fboxsep{0pt}
			\tiny
			\put(1, 25.2){\pgfsetfillopacity{0.8}\colorbox{white}{\framebox(27,4){\pgfsetfillopacity{1}Parked classification}}}
	\end{overpic}}
	\hfill
	\subfloat{%
		\begin{overpic}[width=0.32\textwidth, trim=0cm 0cm 0cm 3cm, clip, frame]{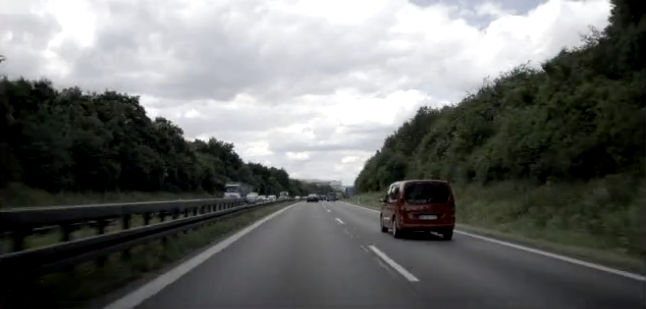}
			\setlength\fboxsep{0pt}
			\tiny
			\put(1, 25.2){\pgfsetfillopacity{0.8}\colorbox{white}{\framebox(26,4){\pgfsetfillopacity{1}Ghost classification}}}
	\end{overpic}}
	\\[-0.26cm]
	\subfloat{%
		\begin{adjustbox}{width=0.32\textwidth, trim=5.5cm 9.46cm 5cm 9.8cm, clip, frame}
			\begin{overpic}[angle=13]{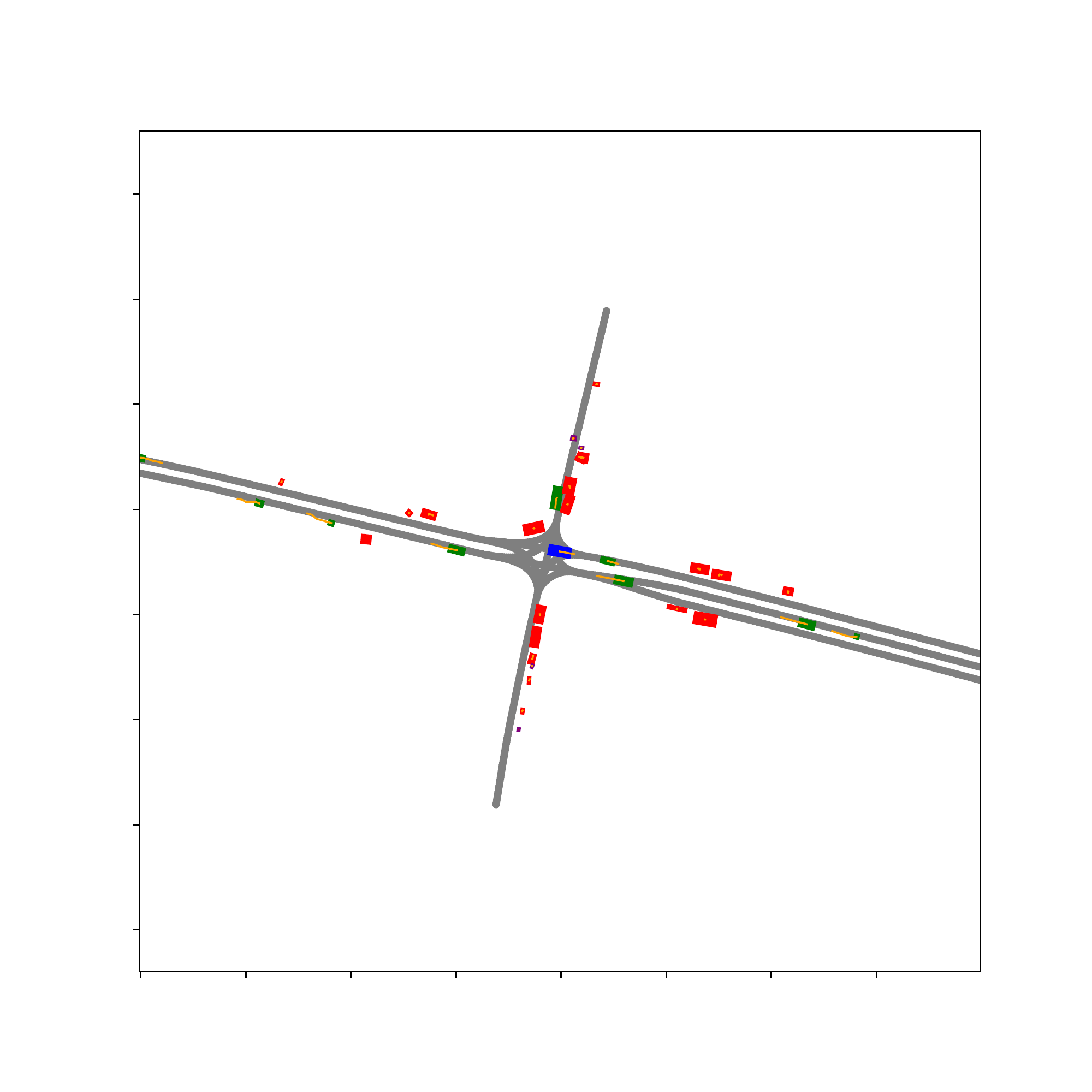}
			\end{overpic}
	\end{adjustbox}}
	\hfill
	\subfloat{%
		\begin{adjustbox}{width=0.32\textwidth, trim=12cm 12.284cm 6cm 10.8cm, clip, frame}
			\begin{overpic}[angle=20]{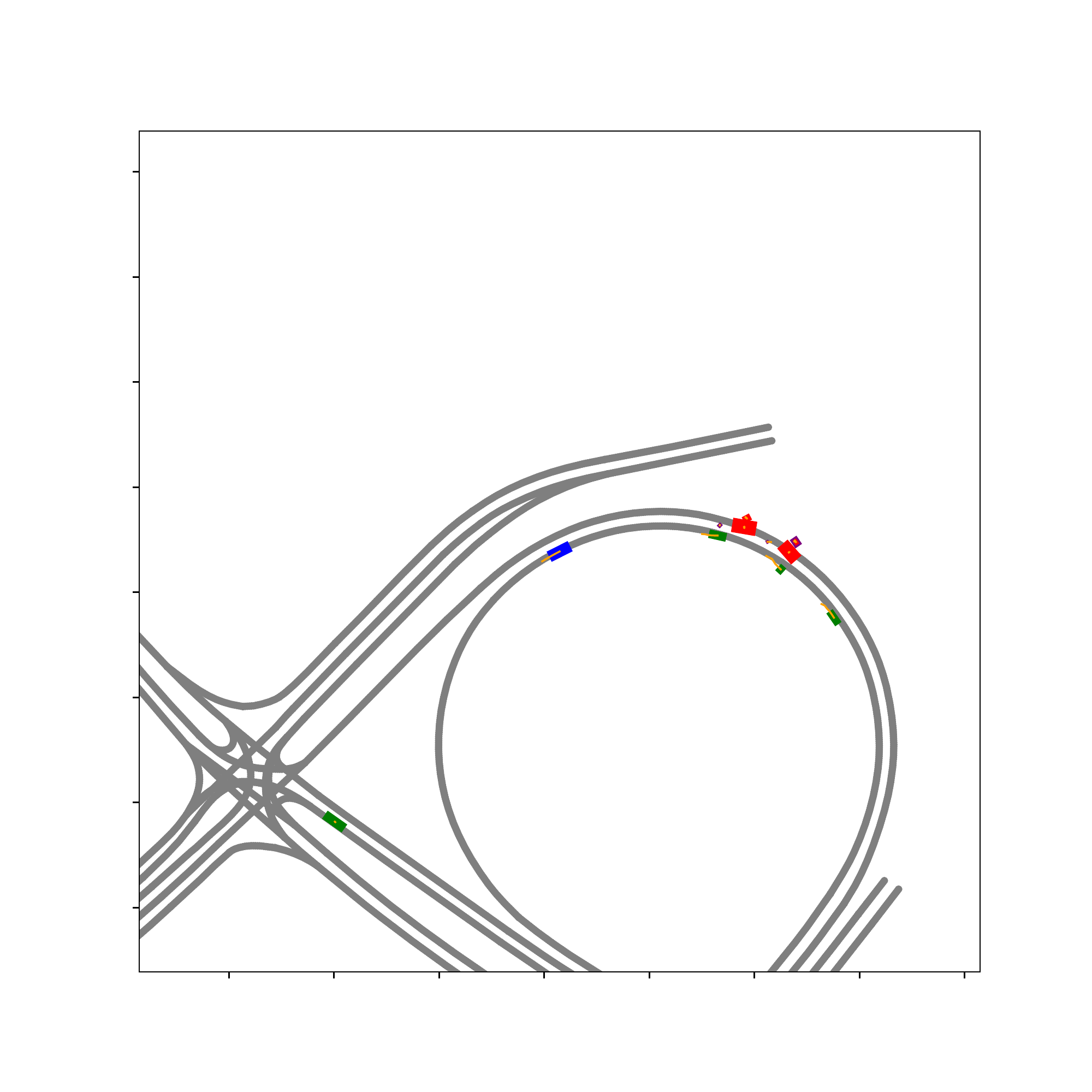}
			\end{overpic}
	\end{adjustbox}}
	\hfill
	\subfloat{%
		\begin{adjustbox}{width=0.32\textwidth, trim=12.1cm 11.96cm 7.1cm 13.17cm, clip, frame}
			\begin{overpic}[angle=321]{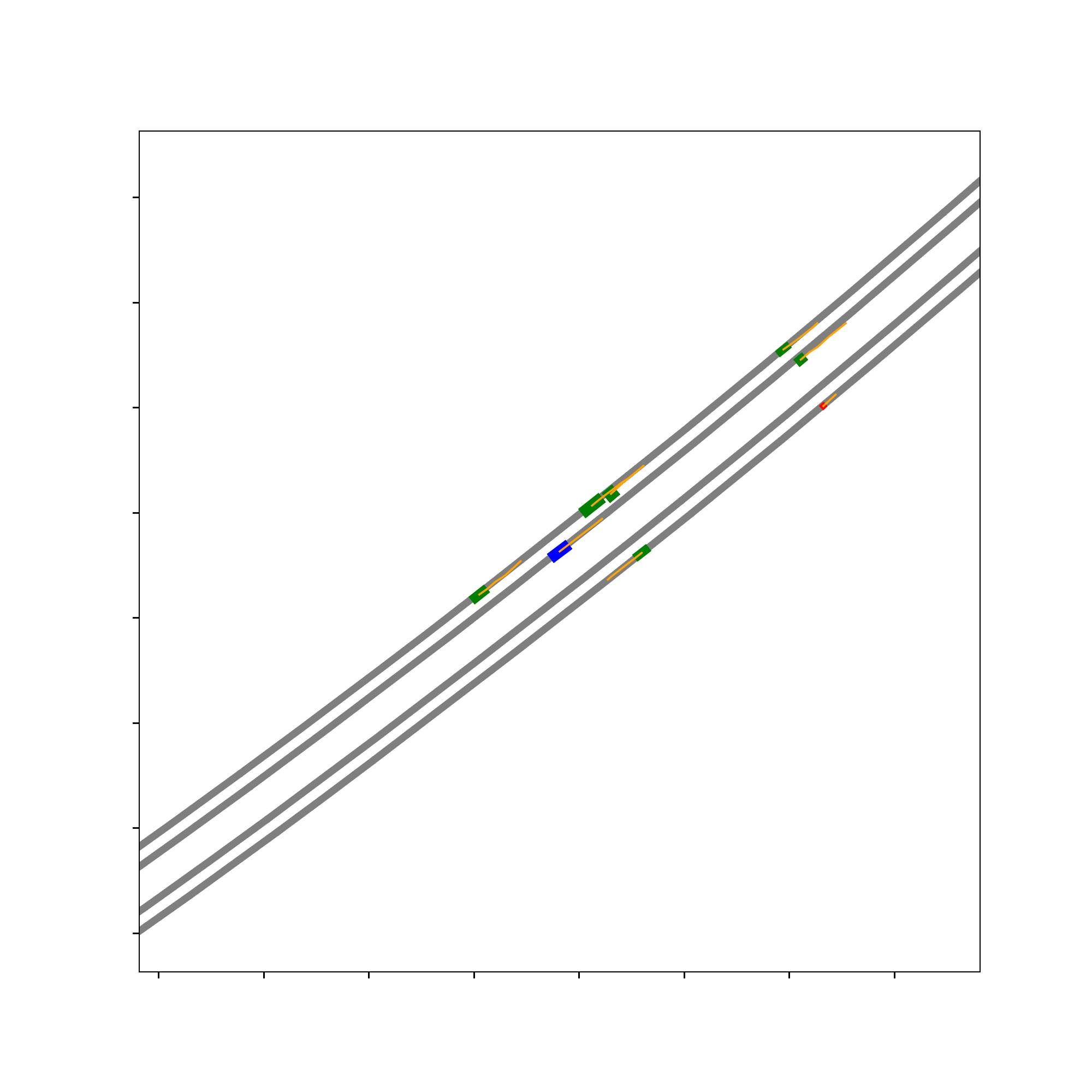}
			\end{overpic}
	\end{adjustbox}}
	\vspace{-0.1cm}
	\caption{Qualitative results of SCENE for the classification of the parked attribute (left and center) and the classification of the ghost attribute (right). The upper row shows the front camera frame, which is one of multiple sensors used by upstream perception, and the lower row renders the corresponding scene with lanes in gray. Bounding boxes of agents are drawn as rectangles, with the past trajectory visualized by an orange line. The color of the rectangle outline indicates the ground-truth label and the fill color indicates the label predicted by our model. Green corresponds to an agent being labeled as non-parked or non-ghost. Red corresponds to an agent being labeled as parked or ghost. All agents are correctly classified as indicated by matching fill and outline colors. A purple outline indicates a missing label. The autonomous vehicle is colored in blue.}
	\label{fig:qualitative_results}
\end{figure*}

\subsection{Ablation Studies}
In two ablation studies we analyze how well our approach can leverage the various sources of information and the effectiveness of our architectural measures.

\begin{table}[!t]
	\scriptsize
	\caption{Context ablation study on the test set}
	\vspace{-0.1cm}
	\label{tab:results_ablation_context}
	\setlength{\tabcolsep}{2.3pt}
	\centering
	\resizebox{1\linewidth}{!}{
		\begin{tabular}{p{0.55cm} p{0.5cm} p{1.10cm} p{0.8cm} p{1.15cm} p{1.15cm} p{1.15cm} p{1.15cm}}
			\toprule
			                                          \multicolumn{3}{c}{Context}                                           & \multirow{2}{*}{\#Params} &                              \multicolumn{2}{c}{Parked}                               &                               \multicolumn{2}{c}{Ghost}                               \\
			\cmidrule(lr){1-3}\cmidrule(lr){5-6}\cmidrule(lr){7-8}
			Agent & Lane                  & Remaining             &                           & F\textsubscript{1} (\%)                   & Acc (\%)                                  & F\textsubscript{1} (\%)                   & Acc (\%)                                  \\ \midrule
			                                                                &                       &                       & $50$k                     & $75.35\scalemath{0.7}{\pm 0.70}$          & $82.96\scalemath{0.7}{\pm 0.54}$          & $80.13\scalemath{0.7}{\pm 1.14}$          & $80.70\scalemath{0.7}{\pm 0.59}$          \\
			\centering \checkmark                                           &                       &                       & $59$k                     & $81.85\scalemath{0.7}{\pm 2.80}$          & $88.31\scalemath{0.7}{\pm 2.00}$          & $79.85\scalemath{0.7}{\pm 0.51}$          & $80.69\scalemath{0.7}{\pm 0.43}$          \\
			\centering \checkmark                                           & \centering \checkmark &                       & $99$k                     & $91.03\scalemath{0.7}{\pm 1.46}$          & $94.28\scalemath{0.7}{\pm 0.89}$          & $\mathbf{80.63}\scalemath{0.7}{\pm 0.63}$ & $81.40\scalemath{0.7}{\pm 0.47}$          \\
			\centering \checkmark                                           & \centering \checkmark & \centering \checkmark & $118$k                    & $\mathbf{91.17}\scalemath{0.7}{\pm 0.71}$ & $\mathbf{94.29}\scalemath{0.7}{\pm 0.46}$ & $80.56\scalemath{0.7}{\pm 0.77}$          & $\mathbf{81.44}\scalemath{0.7}{\pm 0.71}$ \\ \bottomrule
		\end{tabular}}
	\vspace{-0.1cm}
\end{table}
\begin{table}[!t]
	\scriptsize
	\caption{Architectural ablation study on the test set}
	\vspace{-0.1cm}
	\label{tab:results_ablation_architecture}
	\setlength{\tabcolsep}{2.5pt}
	\centering
	\resizebox{1\linewidth}{!}{
		\begin{tabular}{p{0.55cm} p{0.5cm} p{1.10cm} p{0.8cm} p{1.15cm} p{1.15cm} p{1.15cm} p{1.15cm}}
			\toprule
			                                       \multicolumn{3}{c}{Architecture}                                         & \multirow{2}{*}{\#Params} &                              \multicolumn{2}{c}{Parked}                               &                               \multicolumn{2}{c}{Ghost}                               \\
			\cmidrule(lr){1-3}\cmidrule(lr){5-6}\cmidrule(lr){7-8}
			Temp & Res                   & Edge feat\footnotemark &                           & F\textsubscript{1} (\%)                   & Acc (\%)                                  & F\textsubscript{1} (\%)                   & Acc (\%)                                  \\ \midrule
			                                                               &                       &                        & $77$k                     & $89.47\scalemath{0.7}{\pm 1.19}$          & $93.28\scalemath{0.7}{\pm 0.63}$          & $77.42\scalemath{0.7}{\pm 0.96}$          & $78.74\scalemath{0.7}{\pm 0.85}$          \\
			                                                               & \centering \checkmark & \centering \checkmark  & $101$k                    & $90.47\scalemath{0.7}{\pm 0.76}$          & $93.92\scalemath{0.7}{\pm 0.45}$          & $80.26\scalemath{0.7}{\pm 0.55}$          & $80.92\scalemath{0.7}{\pm 0.55}$          \\
			\centering \checkmark                                          &                       & \centering \checkmark  & $102$k                    & $90.81\scalemath{0.7}{\pm 1.17}$          & $94.06\scalemath{0.7}{\pm 0.72}$          & $80.12\scalemath{0.7}{\pm 1.47}$          & $81.02\scalemath{0.7}{\pm 1.26}$          \\
			\centering \checkmark                                          & \centering \checkmark &                        & $111$k                    & $89.06\scalemath{0.7}{\pm 0.72}$          & $92.92\scalemath{0.7}{\pm 0.60}$          & $77.13\scalemath{0.7}{\pm 0.46}$          & $78.71\scalemath{0.7}{\pm 0.23}$          \\
			\centering \checkmark                                          & \centering \checkmark & \centering \checkmark  & $118$k                    & $\mathbf{91.17}\scalemath{0.7}{\pm 0.71}$ & $\mathbf{94.29}\scalemath{0.7}{\pm 0.46}$ & $\mathbf{80.56}\scalemath{0.7}{\pm 0.77}$ & $\mathbf{81.44}\scalemath{0.7}{\pm 0.71}$ \\ \bottomrule
		\end{tabular}}
	\begin{flushleft}
		\footnotemark[2]In contrast to the temporal and residual architectural measures, edge features introduce additional information into the graph.
	\end{flushleft}
    \vspace{-0.3cm}
\end{table}

Table \ref{tab:results_ablation_context} ablates the value of various sources of information coming from the dynamic agents and static infrastructure and the ability of our approach to leverage this information.
Without any context at all, the model performs the classification tasks with the features of \textit{agent} nodes only.
Our experiments show that considering social interactions to nearby agents (agent context), lane information (lane context) and information given by crosswalks, stops and lights (remaining context) can have a strong positive effect on model performance.
The relevance of each contextual aspect differs between tasks.
The remaining context has only a small effect, since much information is implicitly present in the lane model already.
Overall, the results show that incrementally adding further context information improves model performance.
This confirms the value of the additional information as hypothesized in the introduction as well as the capability of the generic model to exploit the provided information.

Table \ref{tab:results_ablation_architecture} ablates the performance of our approach in terms of applied architectural measures.
These measures are the temporal encoding of each agent's trajectory, the residual connections and the inclusion of edge features in the layers of graph convolution.
The results indicate that omitting individual architectural measures decreases model performance compared to applying the full set of measures.
This is particularly noteworthy for the edge features, suggesting a benefit of adding relational information to the graph.
The architecture that combines all measures either excels or comes very close to the best results.
This suggests that the individual architectural measures benefit from each other.

\subsection{Qualitative Results}
Fig. \ref{fig:qualitative_results} shows qualitative results for both tasks.
Color codes are described in the caption of the figure.
In the three examples all agents with available ground-truth label are correctly classified, which is represented by consistent coloring of outline and fill.

The figure on the left shows an urban scenario with vehicles parked on the road side (red) and vehicles driving in the center (green).
Interestingly, the vehicle with white paint inside the paved intersection is parked, which is correctly predicted by our model.

The figure displayed in the center is a rare case where two cars are parked on the left lane of a highway on-ramp.
Again, those are correctly classified by our model.
The prediction is likely supported by the humans nearby, which are detected by the system (purple outline due to no parked label for agents of type human) and provide social context to the parked vehicles.

The figure on the right shows a highway scenario, where all nearby agents besides one are correctly classified as non-ghost (green outline and fill).
The prediction of the one ghost agent (red fill) can be confirmed by its trajectory showing a wrong direction of travel.
This specific detection is probably caused by sensor reflections of a bridge. The corresponding ground-truth label also classifies it as ghost (red outline).

\section{Transferability to other Applications}
As an extension to evaluating the prediction of unknown characteristics of traffic agents, in this section we show that, without any modifications, the use of cascaded layers of graph convolution can be transferred to applications that go beyond the domain of scene understanding.
We therefore apply our methodology for the task of node classification to multiple knowledge graphs of different sizes.
The source code of these experiments, including our graph convolution operator, is publicly available\footnote[3]{Source code: \url{https://github.com/schmidt-ju/scene}}.

\subsection{Datasets}
Evaluation is done on four publicly available heterogeneous knowledge graph datasets, namely AIFB, MUTAG, BGS and AM \cite{Ristoski2016}.
The datasets cover varying graph sizes, ranging from small (AIFB, $8\,285$ nodes) to large (AM, $1\,666\,764$ nodes) \cite{Schlichtkrull2018}.
Given classes for some nodes of a \textit{target} node type, the goal is to correctly classify the classes of masked \textit{target} nodes.

\subsection{Model and Results}
We remove low-degree nodes and initialize the features of each node with a learnable bias vector.
Four layers of HetEdgeGAT are arranged in cascaded form.
The first two layers sequentially update all nodes not of type \textit{target} based on neighboring nodes of the same and of other types.
The second two layers sequentially aggregate information into nodes of type \textit{target} by considering neighboring nodes of other types and of type \textit{target}.
The model is trained full-batch with cross-entropy loss.
Average accuracy and standard deviation for ten runs is reported in Table \ref{tab:results_kg}.
Results of the compared methods are taken from prior publications \cite{Schlichtkrull2018,  Vandewiele2019, Shervashidze2011, Ristoski2019}.

Despite the task being different from predicting unknown characteristics of traffic agents, the results show that our methodology manages to yield state-of-the-art performance for the task of node classification in knowledge graphs.

\begin{table}[!t]
	\scriptsize
	\caption{Accuracy (\%) on the masked nodes of knowledge graphs}
	\vspace{-0.1cm}
	\label{tab:results_kg}
	\setlength{\tabcolsep}{2pt}
	\centering
	\begin{tabularx}{\columnwidth}{Xlllll}
		\toprule
		Dataset & WL \cite{Shervashidze2011}                & RDF2Vec \cite{Ristoski2019}      & Walk Tree \cite{Vandewiele2019}  & R-GCN \cite{Schlichtkrull2018}            & Ours                                      \\ \midrule
		AIFB    & $80.55\scalemath{0.7}{\pm 0.00}$          & $88.88\scalemath{0.7}{\pm 0.00}$ & $89.44\scalemath{0.7}{\pm 2.08}$ & $\mathbf{95.83}\scalemath{0.7}{\pm 0.62}$ & $\mathbf{95.83}\scalemath{0.7}{\pm 1.96}$ \\
		MUTAG   & $\mathbf{80.88}\scalemath{0.7}{\pm 0.00}$ & $67.20\scalemath{0.7}{\pm 1.24}$ & $73.82\scalemath{0.7}{\pm 5.61}$ & $73.23\scalemath{0.7}{\pm 0.48}$          & $75.44\scalemath{0.7}{\pm 2.50}$          \\
		BGS     & $86.20\scalemath{0.7}{\pm 0.00}$          & $87.24\scalemath{0.7}{\pm 0.89}$ & $86.90\scalemath{0.7}{\pm 1.38}$ & $83.10\scalemath{0.7}{\pm 0.80}$          & $\mathbf{92.41}\scalemath{0.7}{\pm 2.72}$ \\
		AM      & $87.37\scalemath{0.7}{\pm 0.00}$          & $88.33\scalemath{0.7}{\pm 0.61}$ & $86.77\scalemath{0.7}{\pm 0.59}$ & $89.29\scalemath{0.7}{\pm 0.35}$          & $\mathbf{90.05}\scalemath{0.7}{\pm 1.07}$ \\ \bottomrule
	\end{tabularx}
	\vspace{-0.2cm}
\end{table}

\section{Conclusion}
This paper proposes a method using cascaded layers of graph convolution in order to predict relevant information from heterogeneous graphs and examines it on the task of reasoning about traffic scenes.
Combining the cascaded layers of graph convolution with our novel way for modeling traffic scenes in heterogeneous graphs results in a generic and extensible method to reason about traffic scenes.
The heterogeneous graph ontology can be extended with additional types or features of nodes and edges.
Our methodology outperforms all task-specific baselines on two diverse tasks.
Furthermore, we compared it to multiple generic state-of-the-art encoders and demonstrated that our method has significant advantages with regard to performance metrics and computational complexity.

The application of our methodology to the task of node classification in knowledge graphs indicates another key property of our methodology:
it is, without any modifications, applicable to areas that go beyond the domain of scene understanding.
By making source code and GNN operator publicly available, we contribute to the progress in this field.

\ifCLASSOPTIONcaptionsoff
  \newpage
\fi

\bibliographystyle{IEEEtran}
\bibliography{Literature}

\enlargethispage{-5in}

\end{document}